\documentclass[sigconf]{acmart}
\usepackage{balance}
\def\BibTeX{{\rm B\kern-.05em{\sc i\kern-.025em b}\kern-.08emT\kern-.1667em\lower.7ex\hbox{E}\kern-.125emX}}

\copyrightyear{2020}
\acmYear{2020}
\setcopyright{acmcopyright}
\acmConference[AIES '20] {2020 AAAI/ACM Conference on AI, Ethics, and Society}{February 7--8, 2020}{New York, NY, USA}
\acmPrice{15.00}
\acmDOI{10.1145/3375627.3375843}
\acmISBN{978-1-4503-7110-0/20/02}

\usepackage{algorithm}
\usepackage[noend]{algorithmic}

\usepackage{mkolar_definitions}
\usepackage{amsmath}
\usepackage{amsthm}
\usepackage{amsbsy}
\usepackage{amsfonts}
\usepackage{multirow}
\usepackage{graphicx}
\usepackage{amssymb}
\usepackage{subfigure}
\usepackage{wrapfig}
\usepackage{float}
\usepackage{color}
\usepackage{url}
\newenvironment{proofsketch}{%
	\renewcommand{\proofname}{Proof Sketch}\proof}{\endproof}

\begin{document}
	
	\fancyhead{}		
	
	\title{Balancing the Tradeoff Between Clustering Value \\ and Interpretability}
	
	\author{Sandhya Saisubramanian}
	\authornote{Both authors contributed equally to this research. \\ This is an extended version of the paper accepted at AIES'20.}
	\affiliation{%
		\institution{University of Massachusetts Amherst}
	}
	\email{saisubramanian@cs.umass.edu}
	
	\author{Sainyam Galhotra}
	\authornotemark[1]
	\affiliation{%
		\institution{University of Massachusetts Amherst}
	}
	\email{sainyam@cs.umass.edu}
	
	\author{Shlomo Zilberstein}
	\affiliation{%
		\institution{University of Massachusetts Amherst}
	}
	\email{shlomo@cs.umass.edu}

	\begin{abstract}
		Graph clustering groups entities -- the vertices of a graph -- based on their similarity, typically using a complex distance function over a large number of features. Successful integration of clustering approaches in automated decision-support systems hinges on the interpretability of the resulting clusters. This paper addresses the problem of generating interpretable clusters, given features of interest that signify interpretability to an end-user, by optimizing interpretability in addition to common clustering objectives. We propose a $\beta$-interpretable clustering algorithm that ensures that at least $\beta$ fraction of nodes in each cluster share the same feature value. The tunable parameter $\beta$ is user-specified. We also present a more efficient algorithm for scenarios with $\beta\!=\!1$ and analyze the theoretical guarantees of the two algorithms. Finally, we empirically demonstrate the benefits of our approaches in generating interpretable clusters using four real-world datasets. The interpretability of the clusters is complemented by generating simple explanations denoting the feature values of the nodes in the clusters, using frequent pattern mining. 
		
	\end{abstract}
	
	
	\maketitle
	
	\section{Introduction}
	% intro to clustering
	Graph clustering is increasingly used as an integral part of automated decision support systems for high-stake applications such as infrastructure development~\cite{hospers2009next}, criminal justice~\cite{aljrees2016criminal}, and health care~\cite{haraty2015enhanced}. Such domains are characterized by high-dimensional data and the goal of clustering is to group these nodes, typically based on similarity over all the features~\cite{jain1999data}. The solution quality of the resulting clusters is measured by the objective value. As the number of features increases, it is increasingly difficult for an end-user to interpret the resulting clusters.
	
\begin{figure*}[t]
		%	\subfigure[Graph]{\includegraphics[width=.75in, height=1 in]{figures/graph-ex.png}\label{graph-ex}}~~
		\subfigure[KC: k-center over all features]{\includegraphics[width=1.3in, height=1.2 in]{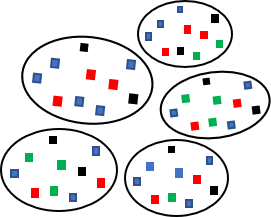}\label{accuracy-ex}} \quad
		\subfigure[$P_F$: Partitions based on FOI ]{\includegraphics[width=1.3in, height=1.2 in]{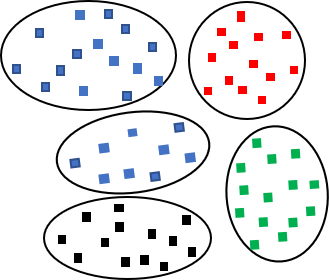}\label{interpretable-ex}} \quad
		\subfigure[$\beta$-KC: Optimized KC \&  interpretability ($\beta=0.8$)]{\includegraphics[width=1.3in, height=1.2in]{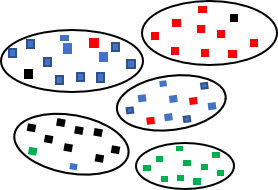}\label{ac-ic-ex}} \quad
		\subfigure[Objective values and corresponding explanations]{\includegraphics[width=2.6in, height=1.5in]{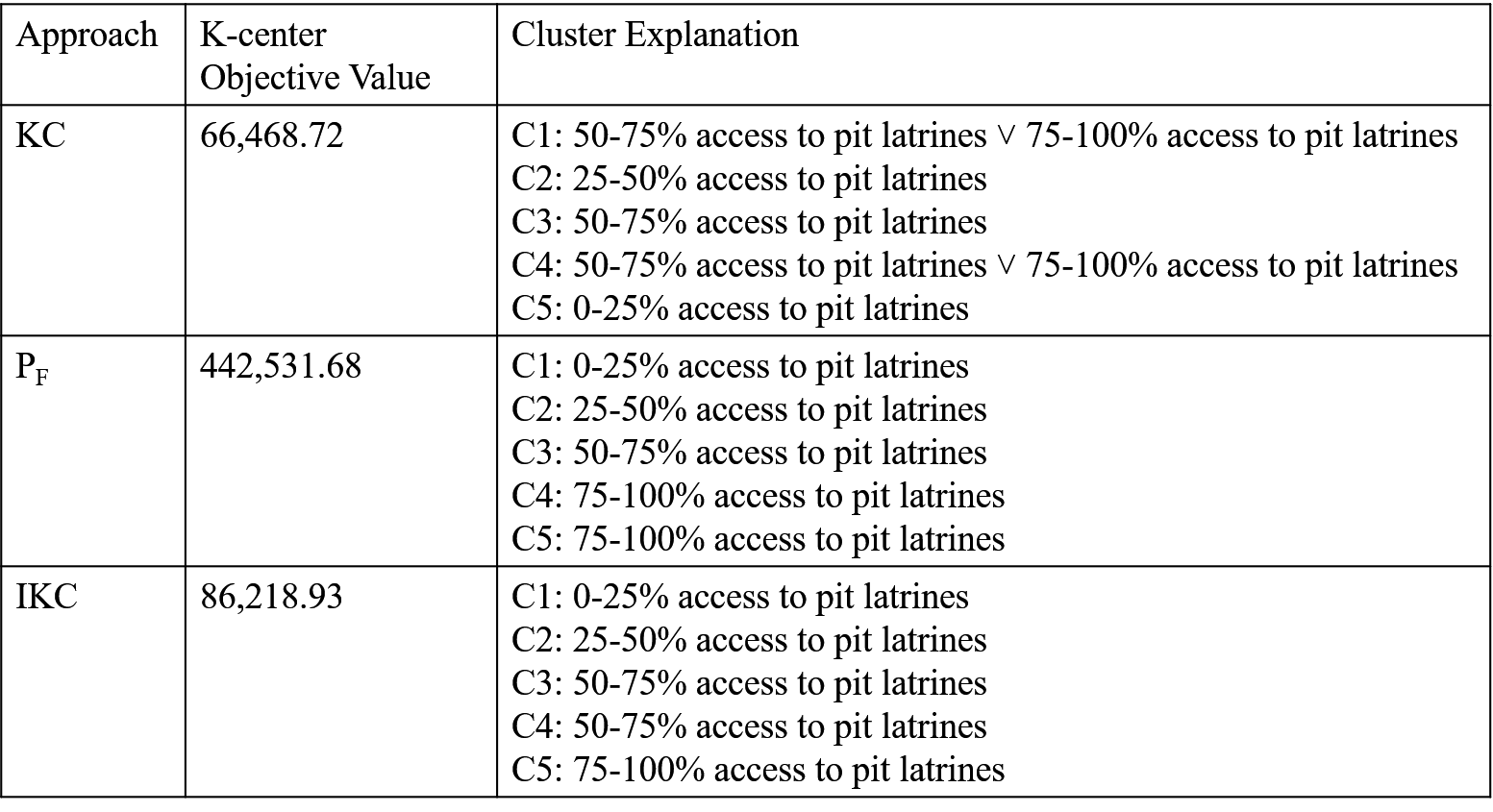}\label{obj-ex}}
		\vspace{-10pt}
		\caption{Illustrative example using the Kenya sanitation data and k=5. The vertices of the graph are the districts with features describing population, sanitation, gender and age demographics, and location. The population \% in each district with pit latrine access is divided into: \{0-25\%, 25-50\%, 50-75\%, 75-100\%\}, indicated by different colors. Interpretability is measured based on access to this basic sanitation. Dominant features in a cluster are generated as labels using frequent pattern mining.}
		\label{fig-example}
		\vspace{-6pt}
	\end{figure*}

	%Example
	For example, consider the problem of clustering districts in Kenya
	to aid decision-making for infrastructure development (Figure~\ref{fig-example}), sanitation in particular~\cite{KenyaVision,sanitation}. Each district is described by features denoting the population, access to basic sanitation, gender and age demographics, and location. The districts in a cluster are typically considered to be indistinguishable and hence may be assigned the same development policies. The similarity of districts for clustering is measured based on all the features. As a result, it is likely that the cluster composition is heterogeneous with respect to the sanitation feature (Figure~\ref{accuracy-ex}). This may significantly affect the decision-maker's ability to infer meaningful patterns, especially due to lack of ground truth, thereby affecting their policy decisions. 
	
	% Interpretable clustering is an open problem.
	Recently, there has been growing interest in interpretable machine learning models~\cite{doshi2017towards,LakkarajuKCL19,rudin2019stop}, mostly focusing on explainable predictive models or interpretable neural networks. There is limited prior research, if any, on improving the interpretability of clusters~\cite{bertsimas2018interpretable,chen2016interpretable}.
	Clustering results are \emph{expected} to be inherently interpretable as the aim of clustering is to group similar nodes together. However, when clustering with a large number of features, interpretability may be diminished since no clear patterns may be easy to recognize for an end-user, as in Figure~\ref{accuracy-ex}. 
	
	Interpretability of the clusters is critical in high-impact domains since decision makers need to understand the solution beyond how the data is grouped into clusters: what characterizes a cluster and how it is different from other clusters. Additionally, the ability of a decision maker to evaluate the system for fairness and identify when to trust the system hinges on the interpretability of the results. 
	%definition of interpretability
	%We employ a domain-independent notion of interpretability, along the same lines as that of explaining black-box classifiers using a set of rules in a well-defined feature space~\cite{LakkarajuKCL19}. 
	In this work, the interpretability of clusters is measured based on the \emph{homogeneity} of nodes in each cluster, with respect to certain predefined feature values of interest (FoI) in the data to the end-user. %Our domain-independent notion of interpretability is simple and effective in helping the user to infer meaningful patterns from the resulting clusters. 
	
	%Competing objectives
	Solution quality of the clusters, denoted by the objective value, and interpretability are often \emph{competing} objectives. % since interpretability typically depends only on a subset of attributes. 
	For example in Figure~\ref{interpretable-ex}, interpretability is optimized in isolation by partitioning the nodes only based on FoI, which significantly affects the solution quality and optimizing for solution quality affects interpretability (Figure~\ref{accuracy-ex}). Reliable decision support requires interpretable clusters, without significantly compromising the solution quality. 
	
	%Our formulation
	In this paper, we study the problem of optimizing for interpretability of clusters, in addition to optimizing the solution quality of centroid-based clustering algorithms such as k-center. We propose a $\beta$-interpretable clustering algorithm that generates clusters such that at least $\beta$ fraction of nodes in each cluster share the same feature value, with respect to FoI. The $\beta$ value is a user-specified input. By adjusting the value of $\beta$, the homogeneity of the nodes in the cluster with respect to FoI can be altered, thus facilitating balancing the trade-off between solution quality and interpretability (Figure~\ref{ac-ic-ex}). We then present a more efficient algorithm to specifically handle settings with $\beta\!=\!1$ and bound the loss in solution quality of centroid-based clustering objectives, when optimizing for interpretability. 
	
	%	Add-ons
	While interpretable clusters are a minimal requirement, it may not be sufficient to guarantee interpretability of the system, due to the cognitive overload for users in understanding the results. Hence, the resulting clusters are complemented by logical combinations of cluster labels as explanations. The feature values of the nodes in the cluster, with respect to FoI, are generated as cluster labels, using frequent pattern mining. In Figure~\ref{obj-ex}, traditional clustering produces longer explanations, which are generally undesirable~\cite{doshi2017towards}, and optimizing for interpretability produces concise explanations. Thus, generating interpretable clusters is crucial for generating concise and useful explanations.

	%Summary of contributions.
	Our primary contributions are: (i) formalizing the problem of interpretable clustering that optimizes for interpretability, in addition to solution quality (Section 2); (ii) presenting two algorithms to achieve interpretable clustering and analyzing their theoretical guarantees (Section 3); and (iii) empirical evaluation of our approaches using four real-world datasets and using frequent pattern mining to generate cluster explanations (Section 4). Our experiments demonstrate the efficiency of our approaches in balancing the trade-off between interpretability and solution quality. The results also show that clusters with different levels of interpretability can be generated by varying $\beta$.

	\section{Problem Formulation}
	Let $V\!=\!\{v_1,v_2,\ldots,v_n\}$ denote a set of $n$ nodes, along with a pairwise distance metric $d\!:\!V\times V\!\rightarrow\!\mathbb{R}$. Let $A^*\!=\!\{A^*_1,\ldots,A^*_m\}$ denote the set of values of  $m$ features where $A^*_i$ refers to the set of values for the i-th feature,  $F^* = \bigtimes \limits_{1
		\le i \le m} A^*_i$ and $\Psi:V \rightarrow F^*$ denote the mapping from nodes to the feature values. Let $H\!=\!G(V,d)$ be a graph where $d:V\!\times V\rightarrow [0,\infty)$  is a metric over $V$. %capturing pairwise relationships such as `do $u,v\in V$ know each other?'. 
	Given a graph instance $H$ and an integer $k$, the goal is to partition $V$ into $k$ disjoint subsets by optimizing an objective function, which results in clusters $\mathcal{C}\!=\!\{C_1,C_2,\ldots, C_k\}$. The objective function ($o$), for a graph $H$ and a set of clusters $\mathcal{C}$, returns an objective value as a real number, $o(H,\mathcal{C})\!\rightarrow\!\mathbb{R}$, which helps compare different clustering techniques. The optimal objective value of an objective function $o$ is denoted by $OPT_o$. $C(u)$ denotes the cluster to which the node $u\!\in\!V$ is assigned. 
	
	The clusters produced by the existing algorithms are often non-trivial and non-intuitive to understand for an end-user due to the complex feature space. Let $A\subseteq A^*$ denote the set of features  in $A^*$ that signify \emph{interpretability} for the user, $F\!=\!\bigtimes \limits_{a \in A} a$ denoting the feature values of interest (FoI). In Figure~\ref{fig-example}, $A$ is the sanitation feature and $F =$\{0-25\%, 25-50\%, 50-75\%, 75-100\%\}, denoting the four feature values of access to basic sanitation. 
	
	\vspace{4pt}
	\noindent{\textbf{Quantifying Interpretability:~}}Interpretability score of a cluster $C$ with respect to a feature value $f\in F$ is denoted by $I_f(C)$ and estimated based on the fraction of the nodes in the cluster that share the feature value, $\forall f \in F$:
	\[ I_f(C) = \frac{\sum_{v\in C}[v_f]}{\vert C \vert}\]
	
	with $[v_f]$ denoting whether the node $v$ satisfies feature value $f$ and $\vert C \vert = \sum_{v\in V}[C(v) = C]$ denoting the total number of nodes in the cluster. Hence, $I_f(C) \in [0,1]$. Given $F$, the interpretability score of a cluster, $I_F(C)\!\in\!(0,1]$, is calculated as \[I_F(C) = \max \limits_{f \in F} I_f(C).\] 
	\begin{definition}
		The interpretability score of a clustering $\mathcal{C}$, given $F$, is denoted by $\mathcal{I}_F(\mathcal{C})$ and is calculated as:\vspace{-2pt}
		\[ \mathcal{I}_F(\mathcal{C}) = \min_{C\in \mathcal{C}} I_F(C). \]
	\end{definition}
	
	\noindent{\textbf{Problem Statement:~}} Given $F$, we aim to create clusters that maximize the interpretability score, $\mathcal{I}_F(\mathcal{C})$, while simultaneously optimizing for solution quality using centroid-based clustering objectives such as k-center. k-center clustering aims to identify k nodes as centers (say $S$, $|S|=k$) and assign each node to the closest cluster center ensuring that the maximum distance of any node from its cluster center is minimized. The objective value is calculated as:
	$$o_{kC}(H,\mathcal{C}) = \max_{v\in V} \min_{s\in S} d(u,s). $$
	
	\begin{definition}
		A clustering $\mathcal{C}$ is $\bm{\beta}$\textbf{-interpretable}, given $F$, if $\mathcal{I}_F(\mathcal{C}) \ge \beta$. That is, each cluster is composed of at least $\beta$ fraction of nodes that share the same feature value.
	\end{definition}

	\begin{definition}
		A clustering $\mathcal{C}$ is \textbf{strongly} interpretable, given $F$, if $\mathcal{I}_F(\mathcal{C}) =1$.
	\end{definition}
	
	%We now analyze the maximum achievable interpretability for a given dataset, $\max_{\mathcal{C}} \mathcal{I}_F(\mathcal{C})$ and identify the upper bound on the value of $\beta$.
	We now analyze the maximum achievable interpretability for a given dataset and identify the upper bound on $\beta$.
	
	\subsection{Optimal upper bound of $\beta$}
	Let $\beta_{max}$ denote the optimal upper bound of $\beta$. Without loss of generality, given a feature value $f\!\in\!{F}$, we assume that there exists at least one node $u\!\in\!V$ that satisfies $f$. When $|F|\!\leq\!k$, with $k$ denoting the number of clusters, a clustering $\mathcal{C}$ can be generated such that $\mathcal{I}_F(\mathcal{C}) =1$. This is achieved by constructing each cluster with the nodes that satisfy the same feature value $f$, and hence $\beta_{max}\!=\!1$.
	
	However, when $|F|\!>k$, there exists no clustering $\mathcal{C}$ with $\mathcal{I}_F(\mathcal{C})\!=1$, since the optimal solution cannot form clusters with nodes satisfying only one feature value of interest. Hence, in such cases, $\beta_{max} < 1$. The optimal value for this case can be estimated as follows: consider the top-k features based on frequency of occurrence in the data and assign the nodes that refer to each of these features to a different clusters. %This results in $k$ clusters, each referring to one of the top-k most frequent features. 
	All the remaining unassigned nodes are then iteratively assigned to the cluster with maximum interpretability score. If multiple clusters have the same interpretability score, the new node is added to the cluster with larger size, since it is less likely to negatively affect the interpretability score.

	In general, the interpretability score of a cluster $C$ is dominated by the feature value satisfied by maximum number of nodes within $C$.  For a given cluster $C$, interpretability can be boosted by either adding more nodes of the majority feature or removing the nodes that are different from the majority feature. If all the nodes that do not represent the majority feature are removed, the interpretability score of $C$ is 1. Using this intuition, we propose algorithms to generate $\beta$-interpretable cluster,  when $\beta \le \beta_{max}$.

	\section{Solution Approach}
	In many applications, the clusters that are considerably homogeneous but not strongly interpretable may still be acceptable since a few outliers do not affect the decision maker's abilities to infer a pattern. For example, if the nodes in a cluster are 90\% homogeneous, the interpretability may not be significantly affected. However, this may help with improving the solution quality of the clusters formed using centroid-based objectives. To that end, we propose an algorithm (Algorithm~\ref{algo:beta}) in which the homogeneity of the nodes in a cluster can be adjusted using a tunable parameter $\beta$. The algorithm identifies $\beta$ interpretable clusters for all values of $\beta \leq \beta_{max}$. We present the algorithms using k-center as the clustering objective. However, it is straightforward to extend the algorithms to any other centroid-based clustering.
	
	\begin{algorithm}[t]
		\caption{$\beta$-Interpretability}
		\label{algo:beta}
		\begin{algorithmic}[1]
			{\scriptsize
				\REQUIRE $G(V,E,d)$, \# clusters $k$, $\beta$, Feature set ${F} \equiv \{f_1,\ldots, f_{|F|}\}$
				\ENSURE Clusters $\mathcal{C}\equiv \{C_1,\ldots, C_k\}$
				\STATE $\mathcal{C}\leftarrow  \texttt{k-center} (V,k) $
				\WHILE{$\mathcal{I}_F(\mathcal{C})$$<\beta$}
				\STATE  $C\leftarrow \arg \min \mathcal{I}_F({C})$
				\STATE \text{majority}$\leftarrow \arg \max_{f_i\in {F}} V_{F_i}\cap C$
				\STATE $S\leftarrow V_{majority}\cap C$
				\IF{$\exists v\in V_{majority}\setminus S$}
				\STATE $\texttt{boost\_majority}(C,\mathcal{C},majority)$
				\ELSE
				\STATE $\texttt{reduce\_minority}(C,\mathcal{C},S)$
				\ENDIF
				\ENDWHILE
				\RETURN $\mathcal{C}$
			}
		\end{algorithmic}
	\end{algorithm}

	The input to Algorithm~\ref{algo:beta} is a graph $G(V,E)$, the parameters $k$ and $\beta$, referring to the number of clusters needed and the interpretability score requirement. First, it initializes a collection of $k$ clusters, $\mathcal{C}$ with the greedy k-center algorithm and optimizes the quality of clusters generated. In order to improve interpretability score, our algorithm iteratively identifies a cluster $C\in \mathcal{C}$ with the least interpretability score and then post-processes it to improve its interpretability scores without considerable loss in the k-center objective. While processing $C$, a feature value $f \in {F}$ associated with maximum number of nodes in $C$ is identified as the `majority' feature value along with a set $S$ corresponding to the collection of nodes that share the majority value. To boost the score of $C$, the fraction of nodes that share the majority feature needs to be increased. We employ the following two operations for this purpose:
	\begin{itemize}
		\item The total number of nodes with majority feature are increased (\texttt{boost\_majority}); and
		\item The nodes that do not correspond to the majority feature value in $C$ are removed from $C$ and re-assigned to other clusters (\texttt{reduce\_minority}). 
	\end{itemize}
	\begin{algorithm}[t]
		\caption{\texttt{boost\_majority}}
		\label{algo:boost}
		\begin{algorithmic}[1]
			{\scriptsize
				\REQUIRE $C,\mathcal{C}$, majority
				\ENSURE 	$C_1,C_2$
				\STATE $C'\leftarrow $ Identify closest cluster based on majority nodes
				\STATE $C_1,C_2\leftarrow identify\_toptwo(C\cup C')$
				\STATE $R\leftarrow C\setminus (C_1\cup C_2) $
				\STATE $R_1,R_2\leftarrow partition(R,|C_1||R|/\theta, |C_2||R|/\theta$),
				where $\theta\!=\!|C_1|+|C_2|$
				\STATE $C_1\leftarrow C_1\cup R_1$, $C_2\leftarrow C_2\cup R_2$
				\RETURN $C_1,C_2$
			}
		\end{algorithmic}
	\end{algorithm}
	
	\begin{algorithm}[t]
		\caption{\texttt{reduce\_minority}}
		\label{algo:reduce}
		\begin{algorithmic}[1]
			{\scriptsize
				\REQUIRE $C,\mathcal{C}$, $S$
				\ENSURE 	$\mathcal{C}$
				\STATE $\gamma\leftarrow |C| - \frac{|S|}{\beta}$
				\STATE $u\leftarrow find\_center(S)$
				\STATE $ X\leftarrow identify\_farthest(C,\gamma,u,S)$
				\STATE Greedily assign all $v\in X$ to the closest cluster $C'\in \mathcal{C}$, ensuring $I_F(C'\cup\{v\})\geq I_F(C')$
				\RETURN $\mathcal{C}$
			}
		\end{algorithmic}
	\end{algorithm}
	\noindent \textbf{\texttt{boost\_majority}.} Outlined in Algorithm~\ref{algo:boost}, this subroutine iterates over the clusters $C\!\in\!\mathcal{C}$ to identify the closest cluster $C'$ that contains the nodes with the `majority feature' and merges $C$ with $C'$ (Line 1,2). It then identifies two different features that have the maximum frequency within the merged cluster and assigns these features to two different clusters $C_1$ and $C_2$ (Line 2). The remaining nodes in the merged cluster are assigned to either of the two clusters such that $C_1$ and $C_2$ have comparable interpretability scores (Line 4,5). 
	
	\vspace{4pt} 
	\noindent \textbf{\texttt{reduce\_minority}.} This subroutine, outlined in Algorithm~\ref{algo:reduce}, identifies the collection of nodes within $C$ that do not have the `majority' feature, which when removed help boost the interpretability score of $C$ (Line 1). Nodes which do not belong to the majority feature and are farthest from the center $u$ are considered for re-assignment (Lines 2,3). Each of farthest node $v\in X$ is then assigned to clusters $C'\in \mathcal{C}$, considered in increasing order of distance from $v$ such that the interpretability score of $C'$ does not reduce below $\beta$ (Line 4). This process of removing nodes from $C$ is performed only when $C$ has the maximum number of nodes present in the data set that share the majority feature.
	
	\begin{remark}
		In some cases, Algorithm~\ref{algo:beta} may converge to a local maxima and may not reach $\beta_{max}$, when the input $\beta =\beta_{max}$. This happens when the feature value being boosted is not one of the feature values in the optimal solution. However, we observe that this is a rare scenario in practice. A detailed algorithm that works in these cases is described in the Appendix.
	\end{remark}

	For cases in which the minimum distance pair identified in Algorithm \ref{algo:boost} belong to same optimal cluster, we bound the loss in k-center objective when using \texttt{boost\_majority}.

	\begin{lemma}
		In each iteration of \texttt{boost\_majority} where the minimum distance pair identified in Algorithm \ref{algo:boost} belong to same optimal cluster,, the k-center objective value worsens by $\alpha OPT_{kC,IC}$, where $\alpha\!\leq\!10$ and $OPT_{kC,IC}$ denotes the optimal k-center objective value of the clusters that achieve maximum interpretability.
	\end{lemma}
	Proof in Appendix.

	When generating clusters with $\beta=1$,  Algorithm~\ref{algo:beta} may take long to converge, especially if the initial k-center based clusters have poor interpretability. We propose a more efficient algorithm for strong interpretability that solves the interpretable clustering problem on each individual features to construct the final solution. 
	
	\subsection{Strong interpretability, $\beta=1$ }
	Algorithm~\ref{algo:beta-one} is a more efficient approach to handle scenarios with $\beta=1$. At a high-level, it identifies the distribution of feature values among $k$ clusters and then quickly generates the clusters. It leverages the property that a clustering $\mathcal{C}$ with $\beta=1$ is characterized by clusters such that all nodes in a cluster share the same feature value. As discussed earlier, $\beta=1$ is achievable only when $|F|\leq k$ and this is an important assumption required for this algorithm.
	
	The first step is to identify a set $\mathcal{S}$ which consists of a $|F|-$tuple of values that sum up to $k$ (Line 2). This set identifies all possible distributions of the different feature values under consideration for interpretability (FoI) among the $k$ clusters. For each value $(s_1,\ldots,s_{|F|})\in \mathcal{S}$, it identifies $s_i$ clusters for nodes with feature $f$. The collection of these $k$ clusters refer to the solution corresponding $(s_1,\ldots,s_{|F|})$ (Lines 3-5). This step generates $|\mathcal{S}|$ collection of k-clusters and the one with minimum k-center objective value is chosen as the final set of clusters (Line 6).
	
	\begin{algorithm}[t]
		\caption{\texttt{strong-interpretability}}
		\label{algo:beta-one}
		\begin{algorithmic}[1]
			{\scriptsize
				\REQUIRE $G(V,E,d)$, \# clusters $k$, Feature set ${F} \equiv \{f_1,\ldots, f_{|F|}\}$
				\ENSURE  Clusters $\mathcal{C}\equiv\{C_1,\ldots, C_k\}$
				\STATE $s_1\ldots s_t \leftarrow 0$
				\STATE $\mathcal{S}=\{(s_1,\ldots, s_{|F|}): s_i> 0, \sum s_i = k\}$
				\FOR{$(s_1\ldots s_t)\in \mathcal{S} $}
				\STATE $\mathcal{C}_{(s_1\ldots s_{|F|})}\leftarrow \cup \texttt{k-center} (V_{f_i},s_i)$
				\ENDFOR
				\STATE $\mathcal{C} \leftarrow identify\_min\_obj(\mathcal{C}_{(s_1\ldots s_{|F|})}, \forall {(s_1\ldots s_{|F|})}\in \mathcal{S})$
				\RETURN $\mathcal{C}$
			}
			
		\end{algorithmic}
	\end{algorithm}
	Algorithm~\ref{algo:beta-one} is capable of generating clusters with high interpretability, without significant loss in the clustering objective value. We now show that the final solution returned by our algorithm is a 2-approximation of the optimal algorithm that generates interpretable clusters and optimizes for the k-center objective.
	\begin{lemma}
		The \texttt{strong-interpretability} clustering algorithm generates $\mathcal{C}$ such that  $\mathcal{I}_F(\mathcal{C})\!=\!1$ and $o_{kc}(H,\mathcal{C})\!=2OPT_{IC,kC}$, where $o_{kC}$ refers to the k-center of objective of $\mathcal{C}$ and $OPT_{IC,kC}$ denotes the optimal k-center objective value of clusters that achieve maximum interpretability.
	\end{lemma}
	\begin{proofsketch}
		Since each cluster $C\in\mathcal{C}$ contains all nodes that share the same feature value,  $\mathcal{I}_{F}(\mathcal{C}) = 1$. Additionally, the optimal solution $OPT_{IC,kC}$ has a distribution of features $(s_1,\ldots,s_{|F|})\in \mathcal{S}$. The solution $\mathcal{C}_{(s_1,\ldots,s_{|F|})}$ is a 2-approximation of $OPT_{IC,kC}$ (following the proof of 2-approximation of greedy algorithm for k-center). Since, the final solution chooses $\mathcal{C}$ that minimizes the k-center objective over all possible clustering in $\mathcal{S}$, it is guaranteed that $\mathcal{C}$ is a 2-approximation of $OPT_{IC,kC}$.
	\end{proofsketch}
	
	\begin{figure*}[t]
		\centering
		\includegraphics[width=7in, height=1.3in]{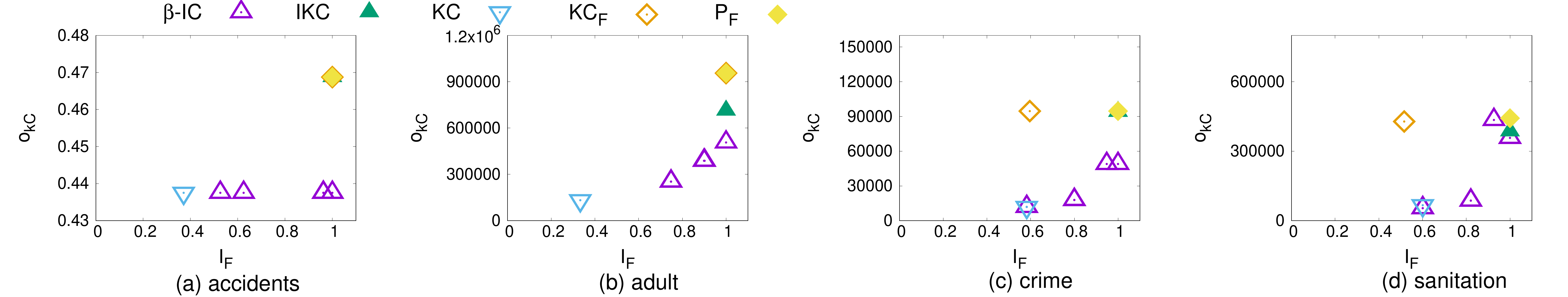}	
		\vspace{-12pt}
		\caption{k-center objective value $o_{kC}$ versus interpretability score $I_F$.}% for varying $\beta$}
		\label{fig-beta}
		%\vspace{-2pt}
	\end{figure*}
	
	\begin{figure*}
		\centering
		\includegraphics[width=7in, height=1.4in]{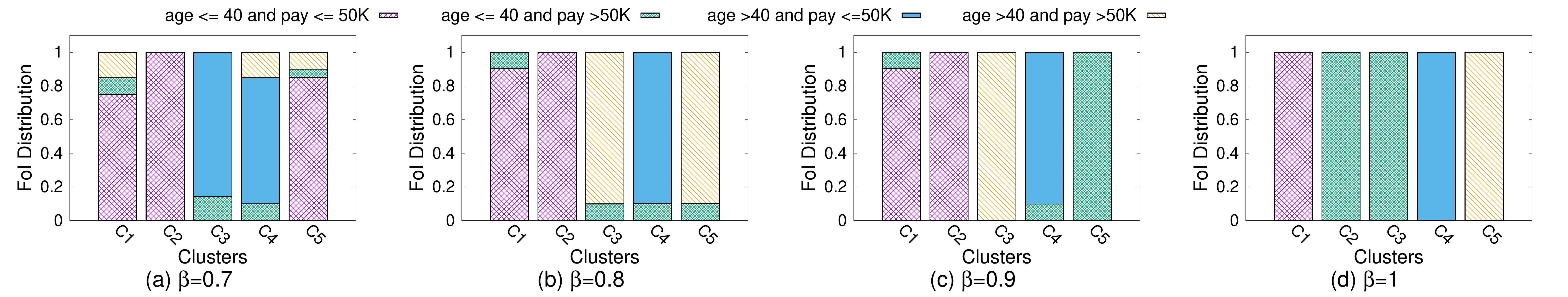}	
		\vspace{-12pt}	
		\caption{Distribution of FoI across clusters with varying $\beta$ on Adult dataset.}
		\label{fig-expbeta}
		%	\vspace{-2pt}
	\end{figure*}

	\section{Experimental Results} We evaluate the efficiency of our approaches based on two metrics: interpretability score of the clustering and the objective value of k-center algorithm. We refer to Algorithm \ref{algo:beta} as $\beta$-IC and Algorithm \ref{algo:beta-one} as $\mathit{IKC}$.
	\vspace{4pt}
	
	\noindent{\textbf{Baselines~}} The results are compared with that of three baselines: 1. k-center clustering over all the features in the data ($\mathit{KC}$); 2. partitioning the dataset into k clusters based on the FoI ($\mathit{P_{F}}$); and 3. k-center clustering over only the features of interest for interpretability ($\mathit{KC_{F}}$). $\mathit{KC}$ and $P_F$ represent extremes of the spectrum, optimizing only for k-center objective or interpretability. $\mathit{KC_F}$ aims to optimize for the distances, ensuring that the nodes with similar features are present close to each other.
	
	\vspace{4pt}
	\noindent{\textbf{Datasets~}} The algorithms are evaluated using four datasets: 1. Kenya sanitation data in which the interpretability is defined over \% population in a district with access to basic sanitation; 2. Kenya traffic accidents data\footnote{https://www.opendata.go.ke/datasets/2011-traffic-incidences-from-desinventar}, whose interpretability is measured based on the accident type; 3. Adult dataset\footnote{https://archive.ics.uci.edu/ml/machine-learning-databases/adult/} in which the  interpretability of the clusters is defined based on the age and income of the population; and 4. Crime data\footnote{http://archive.ics.uci.edu/ml/datasets/communities+and+crime}, with FoI as %the median family income of the communities. 
	the number of violent crimes per 100K population.

	\vspace{4pt}
	\noindent{\textbf{Setup~}} All algorithms are implemented in Python and tested on an Intel i7 computer with 8GB of RAM. In the interest of clarity, we experiment with $\vert F \vert\!=\!4$ for all domains. Due to randomness in the k-center algorithm, the clustering objective behavior of our techniques may not be monotonic. For any given $k\!=\!\theta$, we run the algorithm for different values $k\!=\!\theta'\leq \theta$ and choose the best clustering returned.
	
	%\vspace{4pt}
	%\noindent{\textbf{Solution quality vs. interpretability of clusters~~}}
	\subsection{Solution quality vs. cluster interpretability}
	We first study the trade-off between the k-center objective value and and the interpretability score of the clusters. We vary $\beta \in \{0.5, 0.6, 0.7, 0.8, 0.9, 1.0\}$ for Algorithm~\ref{algo:beta} and compare the results with that of the baselines and that of Algorithm~\ref{algo:beta-one}, with fixed $k=5$. The results in Figure~\ref{fig-beta} show how the k-center objective value may be affected as we form increasingly interpretable clusters using our algorithms. We do not distinguish between the performances with various $\beta$ values, denoted by the purple markers, since the goal is to understand how the algorithm balances the trade-off for any $\beta$ value. We also do not consider $\beta < 0.5$ since that defeats the purpose of optimizing for interpretability. Note that our algorithm supports any value of $\beta \in [0,1]$ as input. 
	
	Approaches that minimize k-center objective and maximize the interpretability, \textbf{lower right} corner of the figure, are desirable. %Clusters formed by $P_F$ and $\mathit{KC_{F}}$ have high interpretability but compromise on the k-center objective value. The greedy k-center algorithm has the best k-center objective but  $\mathcal{I}_F (\mathcal{C})$ is very low. 
	Overall, the baselines either achieve high interpretability with poor k-center objective or low k-center objective with a very low interpretability. Our approach has a better balance between them since the clusters generated by our algorithm have high interpretability, without significant loss in k-center objective.  %Hence, the cost of generating interpretable clusters is lower when using our approach. 
	%Except in a few cases, this trend is fairly consistent for all values of $\beta$ in our experiments. 
	With the increase in $\beta$ values, the k-center objective worsens but the loss in k-center objective is not high and is within a factor of 5 in most cases. The runtime of $\mathit{KC}$ is at most 40 seconds across all datasets and the runtime of our approach is at most 65 seconds across all datasets and all values of $\beta$. This shows that there is no significant overhead in optimizing for both interpretability and solution quality.

	\subsection{Effect of varying $\beta$}
	As discussed above, it is evident that our approaches efficiently balance the trade-off even for higher values of $\beta$. We now study the effects of varying $\beta$ on the cluster composition. Figure~\ref{fig-expbeta} shows the distribution of FoI in each cluster for different values of $\beta$ for the Adult dataset. In the interest of readability, we do not include results for lower values of $\beta$. With the increase in $\beta$, the fraction of majority feature in each cluster grows. For example, the nodes represented by yellow color are merged as $\beta$ is increased from $0.7$ to $0.8$. Similarly, when $\beta$ is increased from $0.8$ to $0.9$, the green colored feature is a minority, which are merged to form a new cluster. Notice that all the green feature nodes are not merged and this process stops as soon as the clusters reach interpretability of $0.9$. However, in the case of strong-interpretability with $\beta=1$, the clusters are homogeneous. In our experiments, the runtime with $\beta=1$ is at most twice as that of $\beta=0.5$ and the runtime of $\mathit{IKC}$ is consistently lower than $\beta$-IC for $\beta=1$. 
	%and our approach ensures uniformity in feature distribution across clusters.
	Similar trends were observed for other domains and the results are included in the Appendix.

	\subsection{Effect of varying $k$}
\begin{figure*}[t]
		\centering
		\includegraphics[width=7in, height=1.3in]{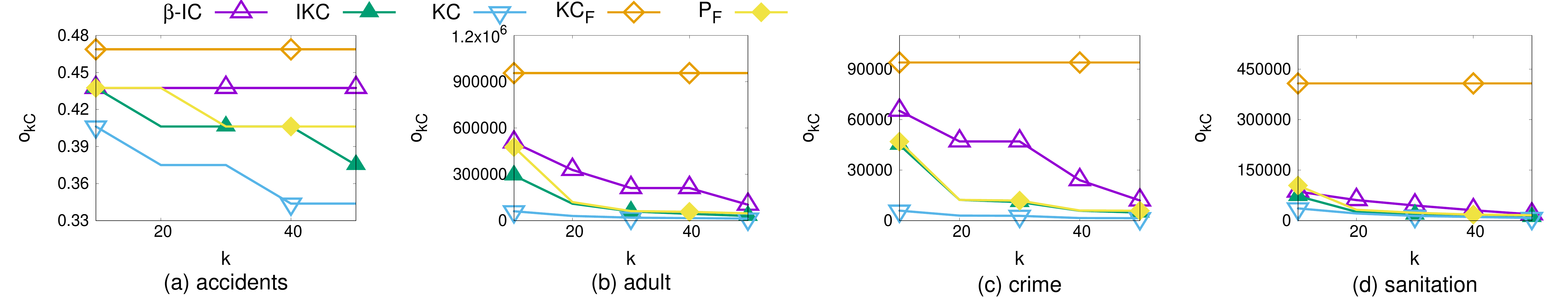}
		\vspace{-12pt}
		\caption{Comparing $o_{kC}$ of various techniques with varying $k$.}
		\label{fig-varyk}
		%\vspace{-2pt}
	\end{figure*}

	To ensure that the trend in the relative performances of the approaches in minimizing the k-center objective are consistent, we experiment with varying the number of clusters $k$, and with fixed $\beta=1$. Figure~\ref{fig-varyk} plots the results of the approaches for $k$ varying from 10 to 50. As expected, the k-center objective value decreases with the increase in $k$ and the relative behavior of all the techniques is consistent.  Our techniques $\beta$-IC and IKC are close to KC across all datasets, while $P_F$ works well for accidents and sanitation datasets only. %In terms of runtime, the additional overhead of $\beta$-IC, $P_F$ and $\mathit{KC}$ is linear with $k$ as opposed to IKC which grows quadratically with $k$. However, all techniques run in less than three hours for all datasets, thereby demonstrating the scalability of our approach.
	All techniques run in less than three hours for all datasets and for all values of $k$ in the experiments, demonstrating the scalability of our approach.
	
	\begin{figure*}
		\centering
		\includegraphics[width=7in, height=1.4in]{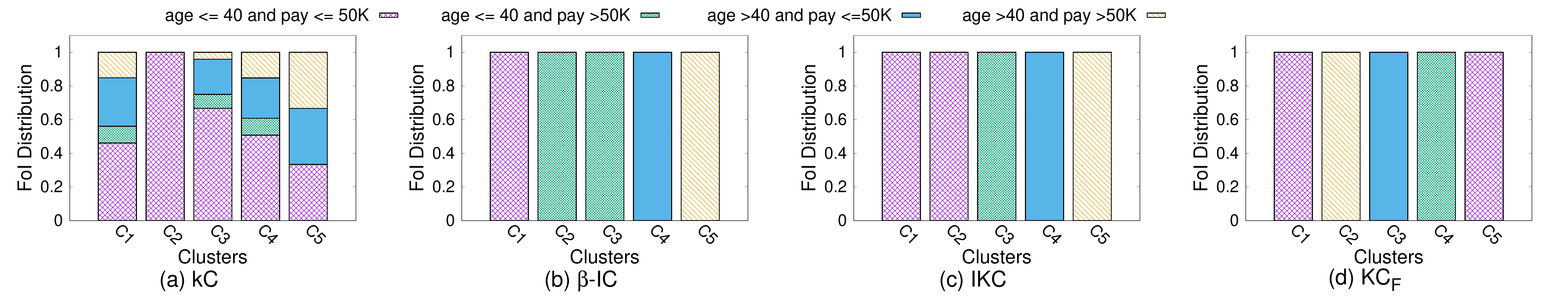}	
		\vspace{-12pt}	
		\caption{Distribution of features of interest (FoI) across clusters for Adult dataset.}
		\label{fig-exptech}
		%	\vspace{-2pt}
	\end{figure*}

	\subsection{Interpretability and Explanation Generation}
	The interpretability of the resulting clusters can be further improved by generating explanations based on the feature values of the nodes in the clusters. Concise and correct explanations based on FoI are possible only when the clusters are homogeneous with respect to FoI. Hence, generating explanations also allows us to understand and compare the performance of different techniques beyond the interpretability scores.
	
	We generate explanations as logical combinations of the feature values of FoI associated with the nodes in each cluster, using frequent pattern mining~\cite{han2007frequent}. This is implemented using Python \texttt{pyfpgrowth} package with a minimum support value of $20\%$ of the cluster size. That is, this approach lists all the feature values that are associated with at least $20\%$ of the nodes in the cluster and $20\%$ is the tolerance for outliers in the cluster. This value can be adjusted depending on the application. Explanations are then generated by a logical OR over these feature values. Figure~\ref{fig-exptech} shows the distribution of explanations across clusters for different techniques on the Adult data, with $\beta=1$. Clusters generated by the $\mathit{KC}$ approach contain a skewed distribution of features across all clusters and are hard to interpret, with respect to FoI. Approaches that focus on interpretability have generated homogeneous clusters with majority of the nodes in a cluster sharing the same feature value. As a result, the generated explanations for these approaches are concise and fairly different across the clusters, thereby improving the interpretability for the decision maker.

	\section{Related Work}
	\noindent{\textbf{Interpretable machine learning~~}} The two main threads of research in interpretable machine learning are generating explanations for black-box models~\cite{abdul2018trends,holzinger2018machine,gunning2017explainable,guidotti2019survey,LakkarajuKCL19} and improving the transparency with interpretable models~\cite{doshi2017towards,rudin2019stop,chen2018looks}. Most of these approaches have been developed for predictive models or for interpretable neural networks and have heavily relied on domain-dependent notions of interpretability~\cite{doshi2017towards}.  We define a domain-independent notion of interpretability and aim to form interpretable clusters, which is critical for high-impact applications~\cite{rudin2019stop}. 
	We argue that generating explanations for clustering requires homogeneous clusters and propose algorithms that improve the interpretability without compromising on the solution quality. 
	
	\vspace{4pt}
	\noindent{\textbf{Clustering with multiple objectives~~}} Prior research on clustering focuses heavily on improving the performance metrics~\cite{aggarwal2010survey,jain1999data,xu2005survey}, such as accuracy, scalability and runtime, but neglect the interpretability aspect. Another thread of work employs soft clustering methods~\cite{chen2016interpretable,greene2005producing} or mixed integer optimization~\cite{bertsimas2018interpretable} to improve interpretability but do not provide any solution guarantees. Constrained clustering~\cite{wagstaff2001constrained}, in which the  pairs of nodes that must belong to the same cluster are enforced as constraints, cannot be used to generate interpretable clusters when $\beta <1$. Another related body of work is the research on multi-objective clustering~\cite{fairlet,law2004multiobjective,jiamthapthaksin2009framework,handl2007evolutionary,bera2019fair} that has been predominantly applied for specific applications and recently for improving fairness. Extending these approaches to our setting is not straightforward since the algorithms are problem-specific. There is limited research on interpretable clustering~\cite{chen2016interpretable} since clusters are expected to be interpretable as they group similar nodes, which is not necessarily the case when dealing with high-dimensional data.

	\section{Conclusion}
	We address the challenge of generating interpretable clustering, while simultaneously optimizing for solution quality of the resulting clusters. We propose an algorithm to generate $\beta$-interpretable clusters, given $\beta$ and the features of interest that signify interpretability to the user. A more efficient algorithm specifically to handle scenarios with $\beta\!=\!1$ is also presented, along with the theoretical guarantees of the two approaches. Our approaches efficiently balance the trade-off between interpretability and solution quality, compared to the baselines. The proposed approach can be extended to handle continuous FoI by treating each interval of continuous values as a discrete value for $\beta$-interpretability. 
	
	We currently target settings in which clustering is performed using centroid-based algorithms. In the future, we aim to expand the range of clustering objectives considered, including hierarchical clustering, and analyze their theoretical guarantees. Using interpretable clustering to identify bias in decision-making is another interesting direction for future research.

\bibliographystyle{ACM-Reference-Format}
\bibliography{ICReferences}

\section{Appendix}
\subsection{Proof of Lemma 5}

\begin{proof}
	In a particular iteration, let the clusters identified to be merged in \texttt{boost\_majority} be $C$ and $C'$ with $d(u,v)$ such that $u\in C$ and $v\in C'$ be the edge with minimum distance. Since  $(u,v)$ are present in the same optimal cluster, then $d(u,v)\leq 2OPT_{IC,kC}$, where $OPT_{IC,kC}$ is the  k-center objective for the optimal clustering that optimizes for IC and k-center.  Hence $OPT_{kC}\leq OPT_{IC,kC}$. The final clusters constructed have nodes from $C$ and $C'$. The maximum pairwise distance between any pair of points $(x,y)$ such that $x\in C$ and $y\in C'$ can be evaluated using triangle inequality. We use the property that the maximum distance between any pair of points within same cluster of radius $r$ is $2r$.
	\begin{eqnarray}
	d(x,y)&\leq& d(x,u)+d(u,v)+d(v,y) \nonumber \\
	&\leq& 4OPT_{kC}+2OPT_{IC,kC}+4OPT_{kC} \nonumber\\
	&\leq & 10OPT_{IC,kC} \nonumber
	\end{eqnarray}
	
	This shows that the pairwise distance between any pair of points within $C\cup C'$ is less than $10OPT_{IC,kC}$. Hence, the final cluster output by \texttt{boost\_majority} will be 10-approximation of the optimal solution in the worst case.
\end{proof}

\subsection{Remark 4}

As discussed in the main paper, Algorithm 1 may converge at a local maxima and never achieve $\beta_{max}$. Even though this is a rare scenario, we present Algorithm \ref{algo:boostinter} as a different subroutine which can be run along with \texttt{boost\_majority} and \texttt{reduce\_minority} subroutines to modify the clusters and identify clustering with  higher interpretability. This subroutine first identifies a feature $f$ which is present as a majority feature in one of the optimal clusters but is not present as a majority in any of the clusters $C\in \mathcal{C}$ (Line 2). Since this feature needs to be a majority, we identify a cluster $C'$ that is closest to $C$ and contains nodes of feature $f$. All nodes with feature $f$ in $C'$ are added to $C$ and then the feature $f$ nodes are boosted to become majority by calling \texttt{boost\_majority} subroutine ensuring that it boosts nodes that belong to feature $f$.

\begin{algorithm}[H]
	\caption{boost\_interpretability}
	\label{algo:boostinter}
	\begin{algorithmic}[1]
		{\scriptsize
			\REQUIRE $C,\mathcal{C}$
			\ENSURE Clusters $\mathcal{C}\equiv \{C_1,\ldots, C_k\}$
			\STATE $F\leftarrow $  Top-k most frequent features
			\STATE $f\leftarrow $ Identify feature that is not majority in $\mathcal{C}$
			\STATE $C'\leftarrow $ Cluster that contains node of feature $f$ and is closest to $C$
			\STATE $C\leftarrow C\cup (C'\cap P_{f})$,  $C'\leftarrow C'\setminus P_{f}$
			\STATE $C\leftarrow$ Boost feature $f$ of $C$
			\RETURN $\mathcal{C}$
		}
	\end{algorithmic}
\end{algorithm} 
\subsection{Additional experiments}
\begin{figure*}[ht!]
	\includegraphics[scale=0.28]{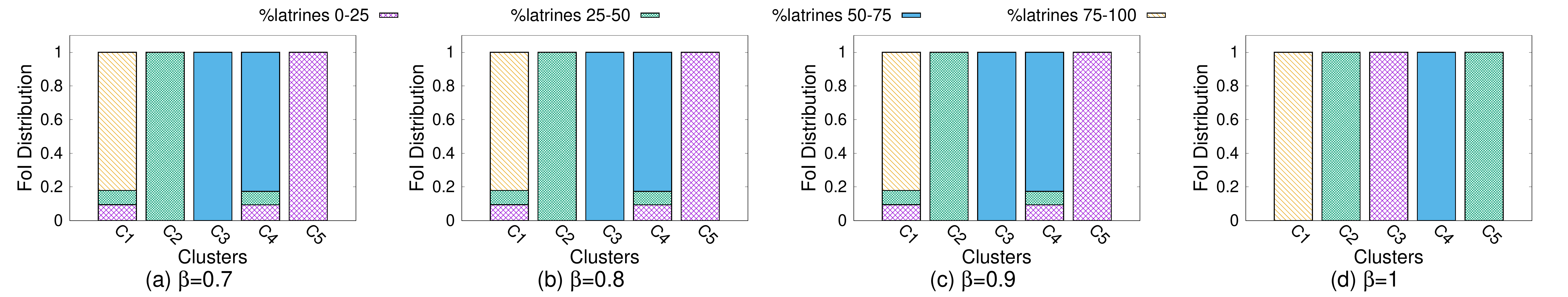}	
	\vspace{-6pt}
	\caption{Results on Sanitation dataset}\label{fig-sanitation}
	\vspace{-6pt}
\end{figure*}

\begin{figure*}[ht!]
	\includegraphics[scale=0.28]{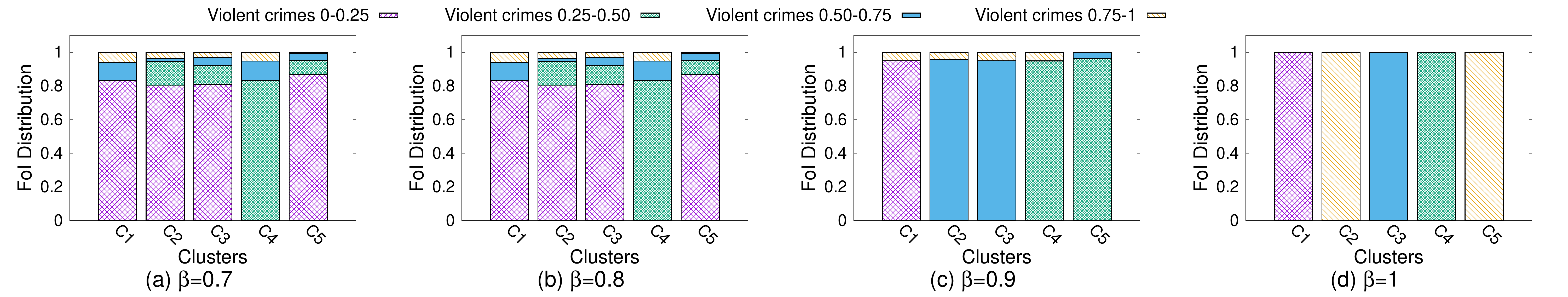}	
	\vspace{-6pt}
	\caption{Results on Crime dataset}\label{fig-crime}
	\vspace{-6pt}
\end{figure*}

\begin{figure*}[ht!]
	\includegraphics[scale=0.28]{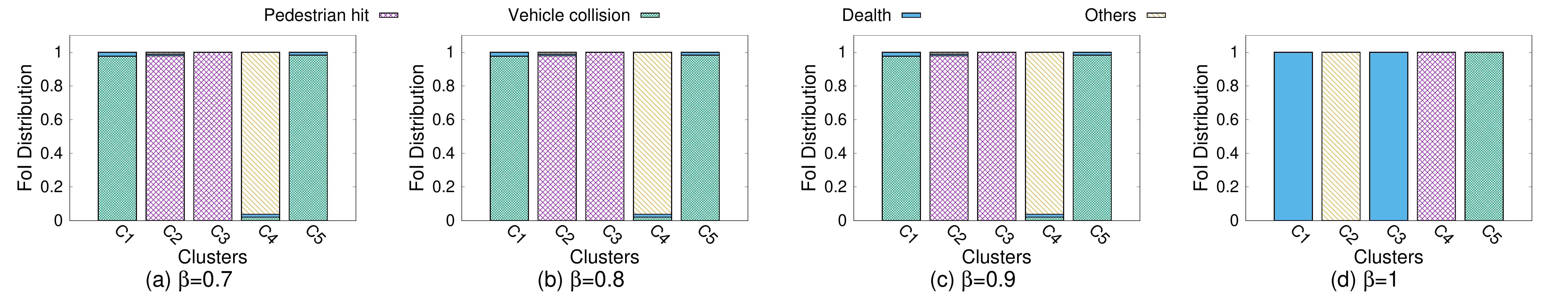}	
	\vspace{-6pt}
	\caption{Results on Accident dataset}\label{fig-accident}
	\vspace{-6pt}
\end{figure*}

As discussed in Experiments section, the distribution of features in different clusters changes with variation of $\beta$. Figure \ref{fig-sanitation}, \ref{fig-crime},  \ref{fig-accident} present the variation in the distribution of FoI across different datasets as $\beta$ is altered. 

%\end{document}		
\end{document}